\documentclass[runningheads,a4paper]{llncs}

\usepackage{graphicx}
\usepackage{textcomp}
\usepackage{listings}
\usepackage{subfiles}
\usepackage{subfigure}
\usepackage{amsmath}
\usepackage{url}
\usepackage{enumitem}  
\usepackage[utf8]{inputenc}
\usepackage{csquotes}
\usepackage{fancyhdr}
\usepackage{float}
\usepackage[T1]{fontenc}
\usepackage[utf8]{inputenc}
\usepackage{authblk}

\begin{document}

\pagestyle{headings}  

\title{Object Detection and Recognition of Swap-Bodies using Camera mounted on a Vehicle}

\titlerunning{Object Detection and Recognition of Swap-bodies using Camera}  

\author{Ebin Zacharias\inst{1}, Didier Stricker\inst{2}, Martin Teuchler\inst{3} \and Kripasindhu Sarkar\inst{4}}

\authorrunning{Zacharias et al.} 
\institute{\email{$^1$ebinzacharias@gmail.com}, \email{$^2$didier.stricker@dfki.de}, \email{$^3$Martin.Teuchler@kamag.com} and \email{$^4$kripasindhu.sarkar@dfki.de}}

\maketitle              

\begin{abstract}
Object detection and identification is a challenging area of computer vision and a fundamental requirement for autonomous cars. This project aims to jointly perform object detection of a swap-body and to find the type of swap-body by reading an ILU code using an efficient optical character recognition (OCR) method. Recent research activities have drastically improved deep learning techniques which proves to enhance the field of computer vision. Collecting enough images for training the model is a critical step towards achieving good results. The data for training were collected from different locations with maximum possible variations and the details are explained. In addition, data augmentation methods applied for training has proved to be effective in improving the performance of the trained model. Training the model achieved good results and the test results are also provided. The final model was tested with images and videos. Finally, this paper also draws attention to some of the major challenges faced during various stages of the project and the possible solutions applied.
\end{abstract}

\section{Introduction}

Computer vision is not a recent topic of research but has a long tradition associated with it. The relevance of research and development in computer vision is significant. Computer vision enables machines to visualize various objects, extract information and understand its surroundings \cite{kendall2019geometry}. Object detection and recognition are the two important tasks of computer vision.

Object detection determines the scope or presence of an object in a given image whereas object recognition identifies the object class to which the object actually belongs to \cite{jalled2016object}. This project aims at training and implementing a deep learning-based object detection model and identifying the text using an optical character recognition (OCR) method. The end goal is to combine the system on an autonomously driven logistic truck in order to improve the efficiency in detecting the type of swap-bodies at a good distance. The logistic yard under consideration consists of 170 different gates for loading and unloading. Containers are located at different locations in the yard and it is taken to these gates by logistic trucks manually. Automating this task improves productivity and efficiency in comparison to manual processes. The final system including object detection and detection of ILU code using OCR technique is elaborated in this paper.

Nevertheless, it is important to note the difficulties in achieving such challenging tasks. The field of computer vision deals with mimicking the complex human visual system to enable a computer to identify various objects as humans do. Besides the complexity of the human visual system, factors such as illumination, perspective variations, occlusion, viewpoint and pose fluctuations also makes object detection and recognition harder \cite{zhao2019object}. Deep learning techniques are emerging to be more powerful in terms of accuracy and performance. The hidden layers in the architecture of deep learning effectively learn different categories from low level to higher levels incrementally. In contrast to traditional methods, the ability to automatically learn high-level features make deep learning techniques firmer \cite{liu2018deep}. 

Various deep learning-based object detection algorithms are available. In the beginning, some of the algorithms were compared and tested in order to get a better understanding of the performance and ease of implementation using Tensorflow. A detailed description of the algorithms compared are discussed in the upcoming sections. The number of images for training is crucial for deep learning models. The larger and diverse the dataset is, the better will be the final trained model \cite{zhao2019object}. Accordingly, high-quality data is gathered and annotated manually using software called LabelImg.

The project is accomplished in two steps. First, the swap-body is detected using a Single Shot MultiBox Detector (SSD), Then the type of detected swap-body is verified by reading the Intermodal Loading Unit (ILU) code using an efficient OCR tool. Even though testing object detection with OCR in videos accompanied various difficulties initially, the final results proved to be a good performing model. The final model is tested using both images and videos under different scenarios. The following sections sequentially explain the different stages of the project.

\section{Object Detection}

Object detection identifies the presence of any object from a predefined category in a given image and outputs the spatial location and extent with a bounding box for every instance of the objects identified \cite{liu2018deep}. There is a wide range of applications for object detection. Face detection in mobile cameras, surveillance systems, detection of a wide variety of objects in autonomous driving and augmented reality are some of the many applications of object detection \cite{jalled2016object}. Furthermore, object detection helps in solving many complex computer vision problems such as scene understanding, segmentation and object tracking \cite{liu2018deep}. 

However, there are limitations in traditional object detection techniques which uses low-level feature descriptors. The lower accuracy levels are one of the major concern \cite{kendall2019geometry}. Notably, the evolution of deep learning techniques has improved the accuracy levels significantly \cite{zhao2019object}. These techniques are powerful and efficient since they automatically learn the feature representations directly from the input data \cite{liu2018deep}.  

The use of Convolutional Neural Networks (CNN) has improved the accuracy of object detection in recent years. Faster Region-based CNN (Faster R-CNN), Region-based Fully Convolutional Network (R-FCN), Region-based CNN (R-CNN), Single Shot Multibox Detector (SSD) and You Only Look Once (YOLO) are some of the CNN based networks which are proved to be good in real-time applications \cite{huang2017speed}. However, there is no consolidated way to choose an appropriate CNN architecture for different applications \cite{huang2017speed}. The selection of different architectures depending on the use case and other factors. The approach for selecting a suitable algorithm for this project is elaborated in the following section.

\subsection{Comparison of object detection algorithms}

Algorithms perform with different accuracy rates and speed. In order to determine the best algorithm suitable for this project, Faster RCNN, YOLOV3 and SSD were trained and the results were compared. YOLOV3 was trained and implemented using the open-source neural network framework called Darknet whereas TensorFlow was used to implement Faster RCNN and SSD. The object detection API from TensorFlow is an open-source and permits easy construction, training and deployment of the model \cite{website2}. Even though python was used for this project, TensorFlow also supports multiple languages such as C++ and R \cite{website2}. Most importantly, the flexible architecture of TensorFlow supports deploying text-based models which is a determining factor in choosing this framework as the project focuses on optical character recognition (OCR) also.

The three algorithms performed well during testing. YOLO V3 was fast but with comparatively lower accuracy rates. In contrast, Faster R-CNN was slower but resulted in better accuracy levels. However, SSD – Inception V2 outperformed YOLOV3 and Faster R-CNN in terms of accuracy and speed respectively. Figure 1 shows a comparison between the three object detection algorithms in terms of speed and accuracy \cite{website1}. Nvidia Jetson developer kit was used for testing the model and based on the computational power available, SSD was a better choice among the three algorithms. Even though Faster RCNN performed better than SSD in detecting small objects, SSD detected large objects with accuracy levels comparable to Faster RCNN. Moreover, the swap-body to be detected is a large object and hence SSD was chosen.  The comparison between accuracy and size of the object is depicted in Fig. 2 \cite{website1}. This project utilizes the SSD architecture using Inception V2 for training the object detector.

{\begin{figure} [h]
\centering
\begin{minipage}{.5\textwidth}

  \centering
  \includegraphics[width=1.1\linewidth]{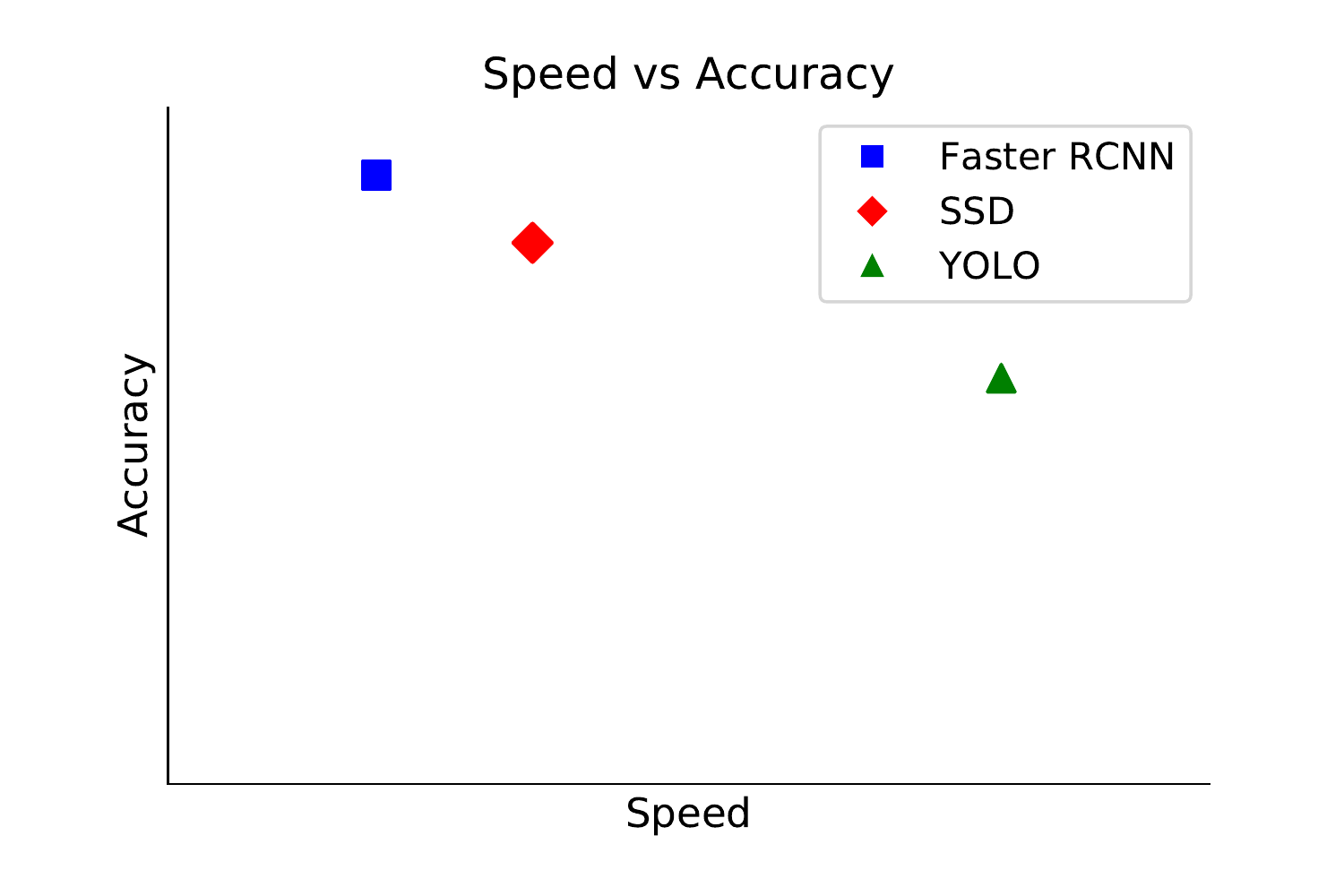}
  \caption{Speed vs Accuracy.}
\end{minipage}%
\begin{minipage}{.5\textwidth}
  \centering
  \includegraphics[width=1.1\linewidth]{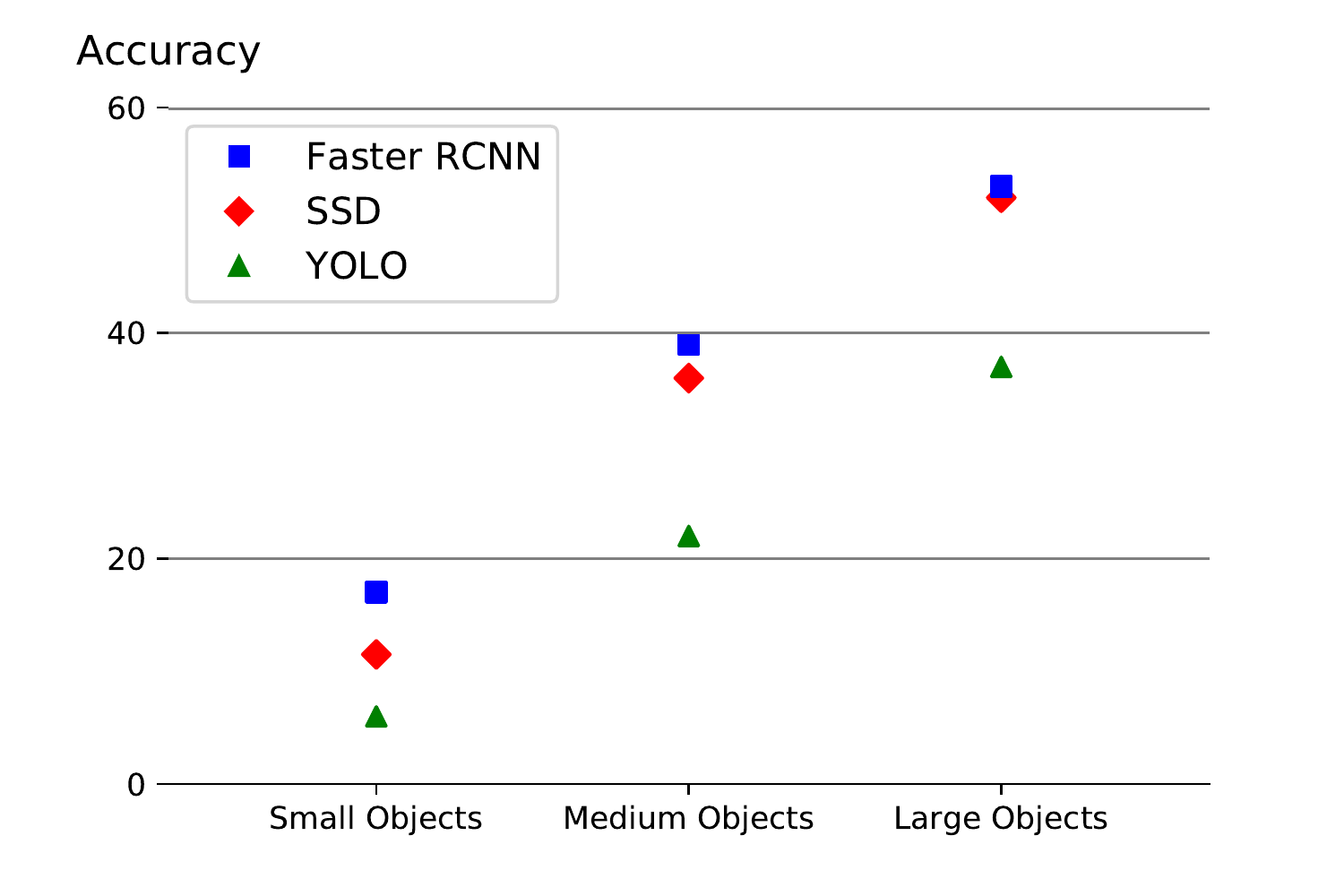}
  \caption{Accuracy vs size of objects.}
\end{minipage}

\end{figure}

\subsection{Methodology}

The project focuses on the detection and identification of swap-bodies using a camera mounted on a logistic truck. A swap-body is a type of freight container used for transportation on roads and rail. It is similar to the ISO containers used for logistic purposes all over the world. However, the concept of swap-bodies is practised in Europe. Swap-bodies are considered to be economical for European Logistics. A swap-body has more capacity as it is scaled to the maximum allowable width and length for European trucks. It also consists of folding legs under the frame which makes easy and efficient transportation \cite{website6}. The truck is responsible for transporting swap-bodies from one location to another within the logistic yard. Based on the dimensions, the swap-bodies in the logistic yard under consideration are categorised into mainly two types: SJSB type and SCSB type of swap bodies. Both the type of swap-bodies are similar in visual appearance and the distinguishing factor is the Intermodal Loading Unit (ILU) for each. 

Intermodal Loading Units are unique numbers printed on the three sides of every swap-body which makes it distinguishable. ILU codes depend on the type of task and the mode of transportation involved. After detecting the swap-body, the ILU code is recognized using an Optical Character Recognition (OCR) pipeline. Further details regarding OCR are discussed in section 3.

\subsubsection{Camera specification}

The selection of a suitable camera is an important step. The ability to function in different outdoor environments, a good resolution and a reasonable price were the initial requirements of the camera. Considering the required image quality, robustness and the environment of the application, an industrial camera was used for this project. The Gigabit Ethernet (GigE) uEye camera from the Imaging Development Systems GmbH was chosen because it is dust-proof, splash-proof and extremely robust. The camera provides greater flexibility as it has Power over Ethernet (PoE) eliminating the requirement of separate cables. Furthermore, the complementary metal-oxide-semiconductor (CMOS) sensor used is extremely light-sensitive and provides maximum precision in colour rendering and better image quality. Table 1 shows the important specification of the camera.

\begin{table}[h]
\centering
\caption{Specification of the camera used.}
\label{my-label}
\begin{tabular}{l|c}
\hline
\multicolumn{1}{l|}{Camera} & {UI-5580CP Rev. 2}        \\ 

\hline

\textit{Sensor type}       & {CMOS }      \\
\hline 
\textit{Frame rate} & {15.0 fps}                  \\
\hline 
\textit{Resolution (H x V)} & {2560 x 1920}   \\
\hline 
\textit{Resolution} & {4.92 MP}    \\
\hline 
\end{tabular}
\end{table}

\subsubsection{Data Collection}

Gathering enough data for training the object detector is the first and most critical step. In general, very large datasets are required for deep learning and it shows a direct proportionality in training a better model \cite{kendall2019geometry}. The amount and quality of images for training are pivotal in achieving good performance and success of the system. The aim was to collect the maximum number of images from two different logistic locations considering the following important requirements:

\begin{enumerate}[label=(\roman*)]
\item The images should differ a lot from one another.
\item The maximum possible angle of views should be included.
\item The images should be of high quality and varying light conditions.
\item The images should include other random objects in the yard.
\item The images should correspond to the different scenarios in which the object detector will be used.
\end{enumerate}
	
The final dataset was gathered in three stages. Some of the initial problems related to false detections were rectified by adding more images of the swap body. Also, more variations of images along with the wrongly detected objects were collected. Images were carefully analysed and with the above-mentioned requirements in mind, a dataset of 1000 images of the swap-body was collected for training the model. 


\subsubsection{Data Annotation}

The relevance of data collected follows with the effective and accurate annotation of the images. Annotation of images, also known as labelling involves assigning bounding boxes for the objects in the image along with tagging the correct labels or classes \cite{website2}. Different software and on-line-services are available for annotating the images. 

This project attempts to train a single class of the object and the class name \enquote{sb\_DB} is used. Image annotations were done manually using the software LabelImg. It is a free and easy-to-use software which also supports YOLO formats \cite{website3}. Consequently, it was easy to perform training with different object detection algorithms at the beginning. The annotations are saved as XML files. Figure 3 shows samples of images annotated using LabelImg. After annotating the images, the data-set is split into training (80\%) and test datasets (20\%). 

\begin{figure}[H]
\centering
\includegraphics[width=0.9\linewidth]{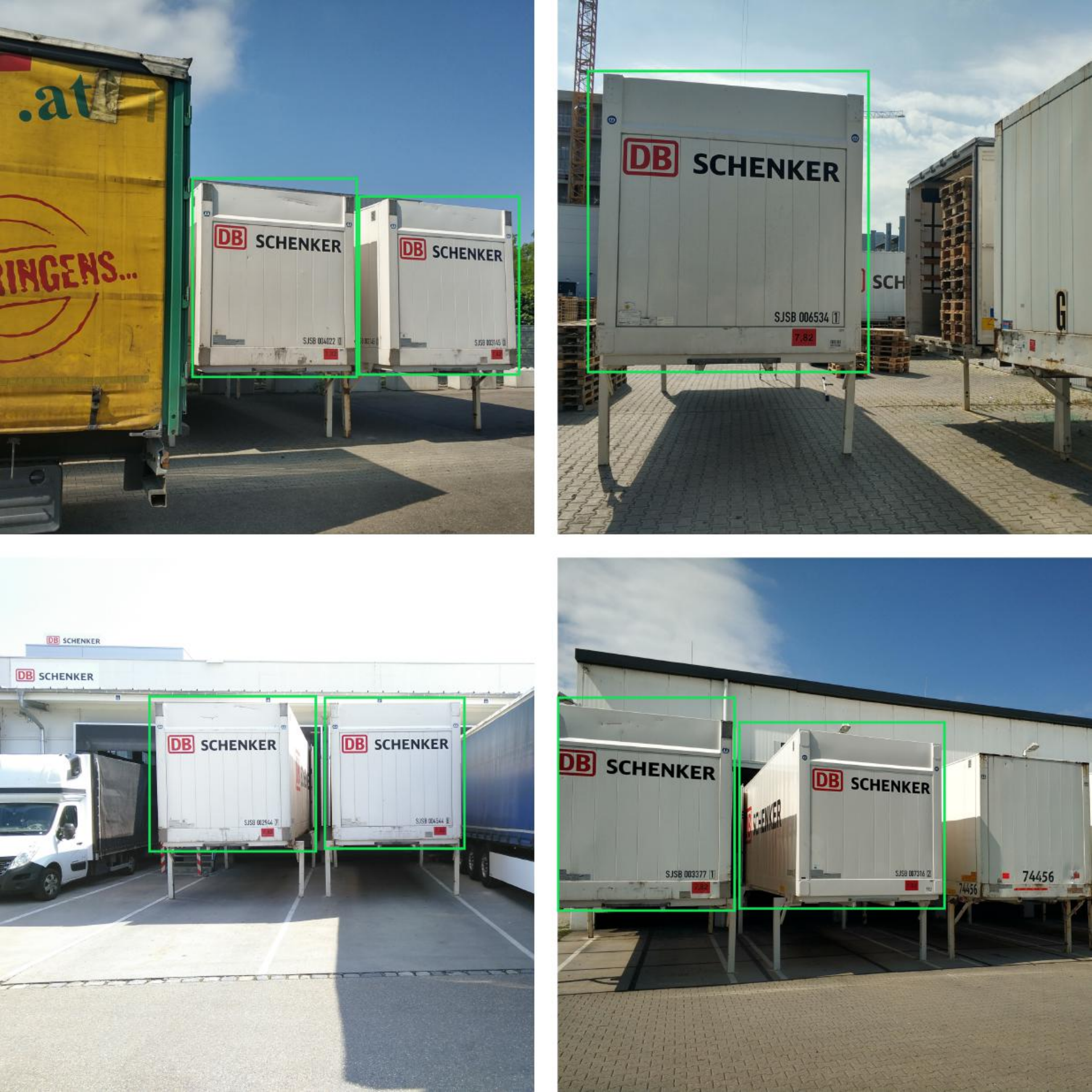}
\caption{A sample of dataset annotated using LabelImg.}
\label{fig:verticalcell}
\end{figure}

\subsubsection{Training}

As discussed earlier, the model is trained using Tensorflow Object Detection API. Tensorflow requires a label map and the image annotations should be in Tensorflow record format (TFRecord). A label map is a text file (*.pbtxt format) which maps each of the labels or classes to integer values \cite{website4}. Label maps are required for both training and detection processes. 

The images labelled using LabelImg is in XML format (*.xml). Prior to converting it into TFRecords (*.records), it is first converted into CSV format (*.csv) \cite{website4}. After conversions, a train.record and test.record files are obtained. The TFRecords are the input data for training the model in Tensorflow.

Training the object detector is a computationally expensive task. Depending on the computational requirements, it can extend from a few hours to even a few days to complete the model training. Hence, the computational power of the hardware used plays a vital role in achieving an efficient training process. Compared to a CPU, training on a GPU is faster and hence more reliable \cite{website4}. For a faster training process,  an N-series virtual machine from Microsoft Azure was utilized for this project. The GPU accelerated platform is advantageous for high-performance computing (HPC) and saves a good amount of time for training. Table 2 shows the specification of the virtual machine used.

\begin{table}[h]
\centering
\caption{Details about the virtual machine used for training the model.}
\label{my-label}
\begin{tabular}{l|c}
\hline
\multicolumn{1}{l|}{Name} & {Standard NC6\_Promo}        \\ 

\hline

\textit{core}       & {6}      \\
\hline 
\textit{RAM} & {56 GB}                  \\
\hline 
\textit{Temporary Storage} & {340 GB}   \\
\hline 
\textit{GPU} & {NVIDIA Tesla K80}    \\
\hline 
\end{tabular}
\end{table}

\subsubsection{Training pipeline configuration}

As mentioned earlier, this project uses a pre-trained model on SSD Inception V2 architecture for training the model. This technique of using a previously trained model for training a new model is known as transfer learning. For fine-tuning the network, the configuration file has to be analyzed and some important changes have to be introduced apart from the hyperparameter tuning. Since the project involves training a single class of object, the number of classes is changed to one in the model configuration file. Similarly, the TFRecord files for training and testing datasets along with the label map files are provided in the configuration file. This completes the initial setup of the model configuration file.

\subsubsection{Hyperparameter tuning}

Hyperparameters are decisive in the network structure and also influences the way a network is trained. An efficient tuning of the hyperparameters before the training process can influence the overall model substantially \cite{website4} \cite{website5}. During the entire project, prime importance was given for attaining the best hyperparameter values. Based on various literature reviews and trial and error methods, the following hyperparameters were altered which resulted in the best performance of the object detection model.

\paragraph{Dropout rate:} Dropout is a method to prevent overfitting. This regularisation technique randomly turns off the neurons in the network during the training process subsequently making the model more resilient \cite{website5}. For this project, dropout was used by setting the probability for dropout to be 0.8.

\paragraph{Batch size:} Batch size determines the number of images given to the network before updating the weights \cite{website5}. The batch size depends on the available memory of the system used and influences the overall training duration. Different batch sizes starting from 2,8,16 and 32 were used during experimentation. Considering the computational power of the system and the training performance, a batch size of 32 was used for the final training. 

\paragraph{Learning Rate:} Learning rate determines the rate at which the parameters of the network are updated. Learning rate is an important hyperparameter as it controls the rate at which the model learns while training \cite{website5}. A very large learning rate can cause overshooting. On the other hand, a small learning rate increases the learning duration. However, it is preferable to start with a smaller value for better convergence \cite{website5}. Various learning rate values were examined and a learning rate value of 0.0001 was considered as the optimal value for the final model training. 

Furthermore, data augmentation techniques were used to make the model more robust and generalize better. Data augmentation enlarges the dataset by artificially transforming the training image. The technique increases the dataset and gives more unique and varied images for training and has an enormous influence in improving the overall object detection accuracy \cite{liu2016ssd}. Since the object to be detected was considerably large, the only geometric distortion used was random cropping. However, more photometric distortions such as random brightness, contrast, hue and saturation were applied.

The goal of the training process was to achieve minimum total loss and also evaluating the model at the same time for the test datasets in order to prevent overfitting. In order to attain the best possible model and also to study how the learning process progresses, a total of 35000 steps were performed. Tensorboard was used for continuously monitoring the training process. 

\subsubsection{Tensorboard: Training process monitoring}

Continuous monitoring of the training job, as it progresses is certainly a critical step. Monitoring the training process is a crucial step as it prevents the model from overfitting or underfitting. Moreover, comparing the loss graphs suggests how good the training is, as it progresses. A graph comparing training and validation loss is shown in Fig. 4. Despite the reduction in training loss, it is clear from the graph that validation loss gradually increases and the gap widens after 15000 training steps. This is a clear indication of overfitting. Hence, the model at 16000 training steps with a training loss of 1.7 was saved and the inference graph was generated. The object detection is executed using the trained inference graph which is generated at the end of the training.

\begin{figure}[H]
  \centering
  \begin{minipage}[b]{0.49\textwidth}
    \includegraphics[width=\textwidth]{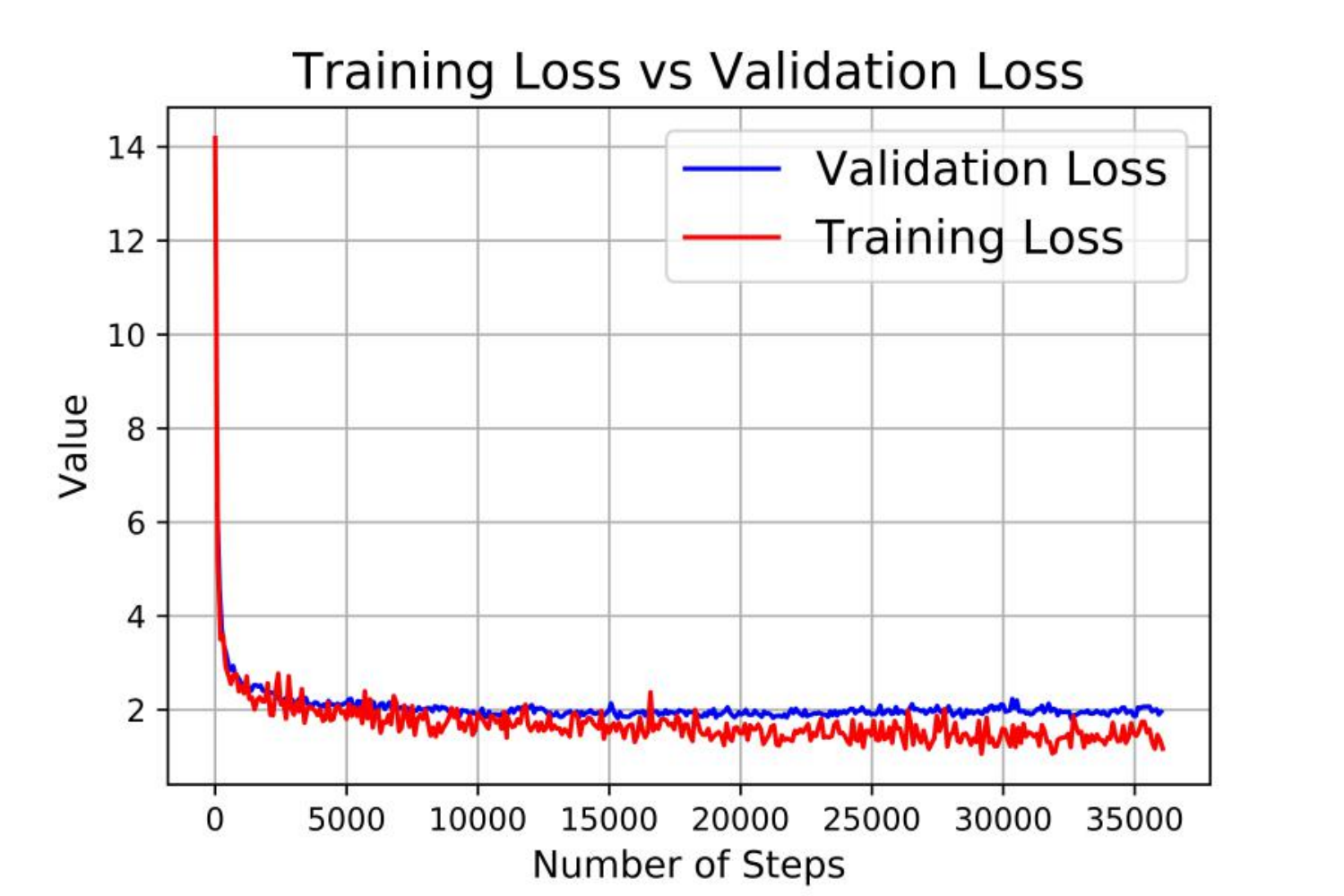}
    \caption{Training loss v/s Validation loss.}
  \end{minipage}
  \hfill
  \begin{minipage}[b]{0.49\textwidth} 
    \includegraphics[width=\textwidth]{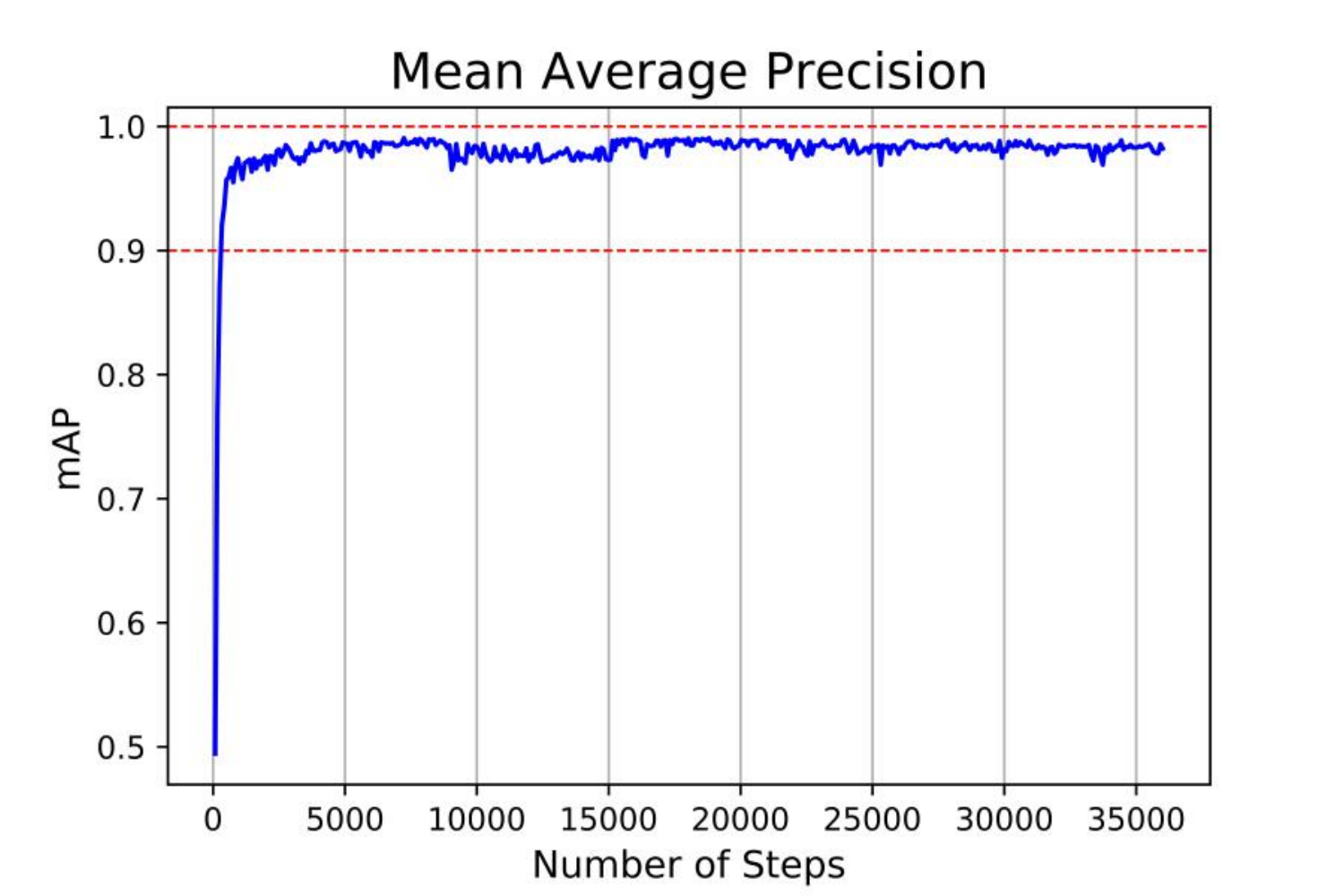}
    \caption{Mean Average Precision (mAP).}
  \end{minipage}
\end{figure}


Tensorboard is such a feature from Tensorflow which enables visualizing different metrics and consequent tracking of the training process. The interface of Tensorboard consists of different tabs such as scalars, graphs, histogram, distribution and projector. Monitoring the scalar values is highly recommended as it gives necessary information about the model training. The tab graph displays the model used for training. A mean average precision of 0.9906 was obtained. Figure 5 displays the graph of the mAP for the entire training process. Clearly, the accuracy rates are higher between 15000 to 17000 steps. 


\subsubsection{Testing and Results}

Testing the accuracy and real-time performance of the trained model is important. It also gives an idea of how robust the model is by evaluating the performance in different situations. Testing was performed at various stages of this project and it helped in rectifying the problems. Since the logistic consists of many other swap bodies which are similar in shape and appearance, it was initially a challenge to avoid wrong predictions made by the model. Another challenge was to differentiate between a trailer and a swap body. As a solution to the latter, images were annotated including the legs of the swap body. The model was trained again with the newly annotated images but the results were not improved substantially. However, both the issues were solved when the model was trained using more number of high-quality images of the swap body. In addition, the increased variations in the final dataset used significantly improved the performance of the final model.

The final inference graph was used to implement object detection in TensorFlow. Testing was performed using the Jetson TX2 developer kit from Nvidia. Random images were taken from the logistic yard and were tested. The earlier problems of false detections were completely solved and the model performed really well in all complex situations. Samples of images tested with the final trained model are shown in Fig. 6. Furthermore, similar results were achieved when the model was tested on videos. Videos were captured from a vehicle moving at different speeds (10 to 30 km/h) in the logistic yard. The model performed remarkably in all the tested driving conditions. Different scenarios were tested and the model performed without errors.

\begin{figure}[H]
\centering
\includegraphics[width=0.9\linewidth]{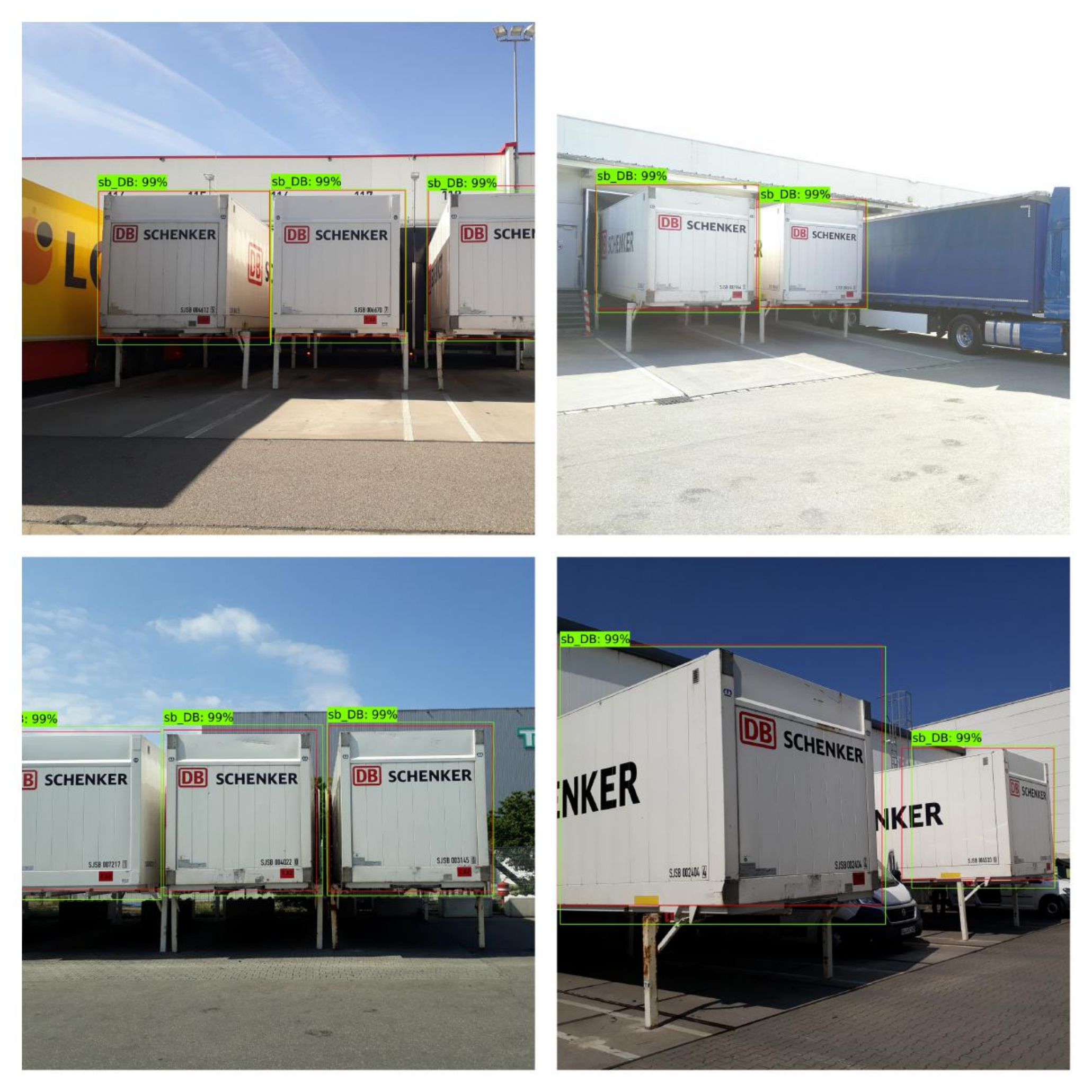}
\caption{Test results of the trained model along with ground truth (red boxes)}
\label{fig:verticalcell}
\end{figure}

\section{Optical Character Recognition (OCR)}

Optical character recognition abbreviated as OCR involves extracting text from images \cite{sun2018review}. OCR is one of the most challenging computer vision tasks. OCR applications vary from well structured, printed texts such as PDF documents to random texts in natural scenes such as reading number plates and street signs. Evidently, the latter is more challenging to achieve, considering the sparse text density, unstructured and noisy text attributes \cite{sun2018review}. This project targets a similar approach to find text in outdoor conditions. After detecting the swap-body,  the detected bounding boxes are used as an input to find the corresponding ILU code.

The goal is to recognize the ILU code printed on the swap-bodies using an efficient OCR technique. The proposed approach in understanding the text progresses as a two-step process. First, instances of texts are detected using a powerful text detector followed by recognizing the detected text using an OCR engine. 

\subsection{Methodology}

Text detection is the first step and plays an important role in the overall performance of the OCR engine. The focus lies in isolating text from its backgrounds. However, distinguishing text from natural scenes is challenging as it is subject to noise and the texts are not well-defined \cite{choudhary2018text}. Deep learning techniques have improved the efficiency and accuracy of text detection in natural images \cite{sun2018review}.

\subsubsection{Effective and Accurate Scene Text (EAST) detector}

Considering the above-mentioned difficulties in scene text detection, for better performances EAST detector based on a fully convolutional network (FCN) is used for identifying instances of text in the image \cite{zhou2017east}. The main reasons to use EAST detector are its efficiency in detecting texts in natural scenes, robustness and ease of implementation using OpenCV. Furthermore, EAST detector accurately predicts the presence of texts in natural scene images along with their geometries \cite{zhou2017east}.

The previously detected bounding boxes for the swap-bodies are the input to the EAST detector. The ILU codes are printed on the lower sides of the swap body. The proposed approach in this project aimed at preventing other random texts from the input to the EAST detector. In order to reduce the complexity and thereby improving the speed of text detection, the bounding box is divided into two halves and only the lower half is considered as the input for text detection. 

In addition to that, EAST detector requires the input size to be in multiples of 32 and the image is resized accordingly. EAST detector outputs bounding boxes around the texts along with the confidence scores. Hence, regions with weaker presence of texts can be ignored by assigning a threshold value based on the confidence scores \cite{zhou2017east}.

\subsubsection{Text Recognition using Tesseract V4}

Text recognition involves understanding and extracting the actual textual content. The detected and localized bounding box coordinates of the text area are taken for recognizing the actual text. The text region of interest (ROI) are passed to an open-sourced OCR engine called Tesseract. This completes the OCR pipeline and it is implemented using OpenCV and Python.

The advancements in deep learning have favoured character recognition also. The latest version Tesseract V4 with LSTM (Long Short-Memory), uses a deep learning-based model which improved the accuracy of OCR results significantly \cite{ramesh2018improving}. There are 3 main important flags that are to be considered while using Tesseract V4. They are:

\begin{enumerate}[label=(\roman*)]
\item Language: Tesseract V4 supports a large variety of languages and for this project, the language of detection is set to English.

\item OCR Engine Mode: OCR engine mode determines which type of algorithm to be used for text recognition by Tesseract. There are 4 different modes and the project utilizes neural nets LSTM only mode as the OCR engine. 

\item Page Segmentation Mode: Page segmentation mode plays a dramatic role in the accuracy of OCR results. There are 13 available page segmentation modes. After a thorough analysis of all the different modes, the mode in which the image was considered as a single line of text was considered for this project. Besides good accuracy rates, the page segmentation mode mentioned was consistent in various test cases. 
\end{enumerate}

\subsubsection{Testing and Results}

In the initial phase, the complete OCR pipeline consisting of text detection using EAST detector and the recognition of detected text using Tesseract V4 was tested for different images. It helped in optimizing the overall OCR output by varying the page segmentation modes. Tesseract performed really well when tested with images and the outputs were accurate. Figure 7 shows the output of detected text boxes using EAST detector. The detected text boxes are then given to Tesseract V4 for identifying the real text. In order to identify only the ILU code, only the text with the strings \enquote{SJSB} or \enquote{SCSB} were considered. A sample of detected ILU code using Tesseract V4 can be seen in fig. 8. The approach for the final system is explained in the following section. For better robustness, the minimum confidence threshold was set to 0.99 and the test results were excellent.

\begin{figure}[H]
  \centering
  \begin{minipage}[b]{0.45\textwidth}
    \includegraphics[width=\textwidth]{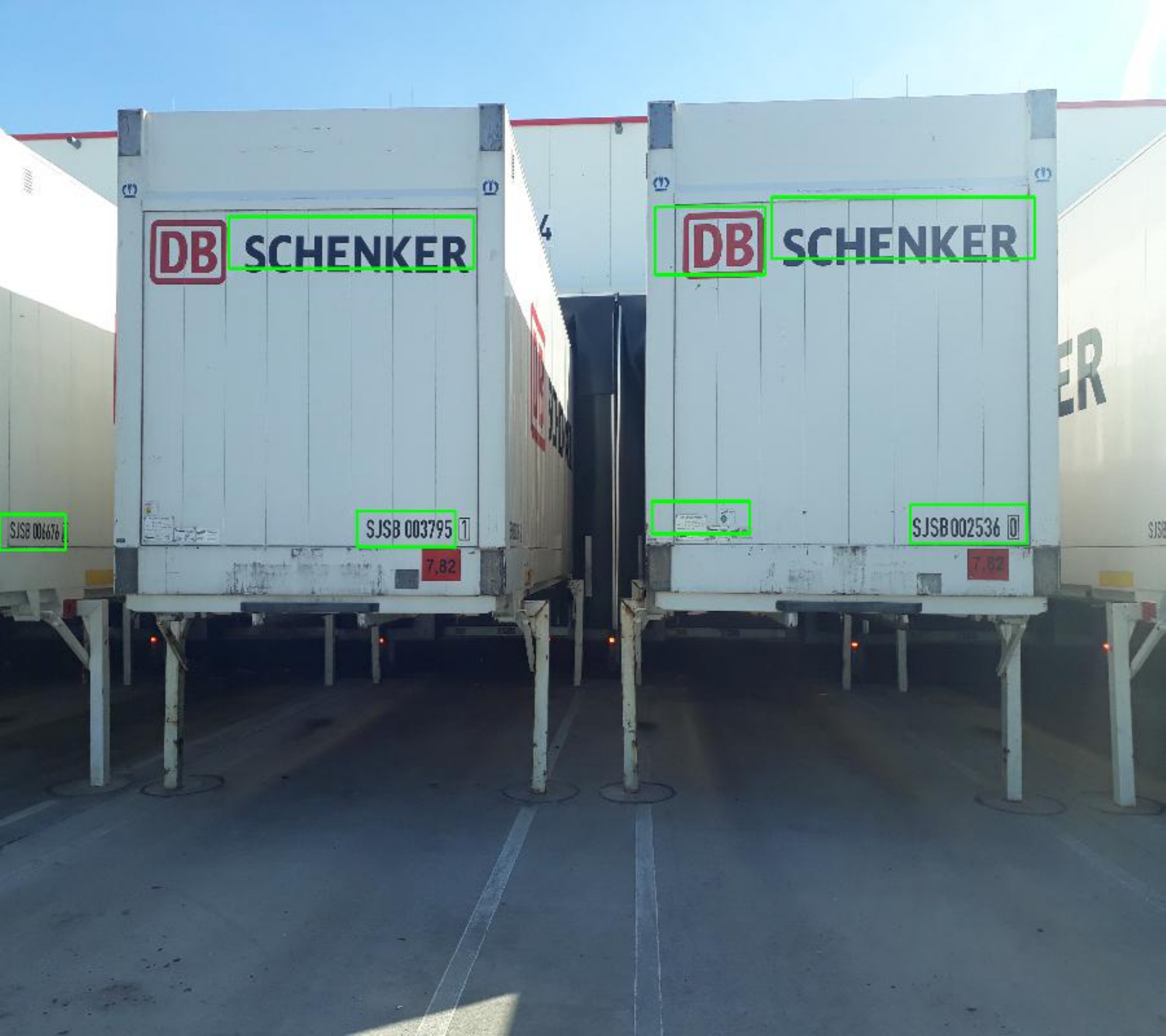}
    \caption{Detected text boxes using EAST detector.}
  \end{minipage}
  \hfill
  \begin{minipage}[b]{0.45\textwidth}
    \includegraphics[width=\textwidth]{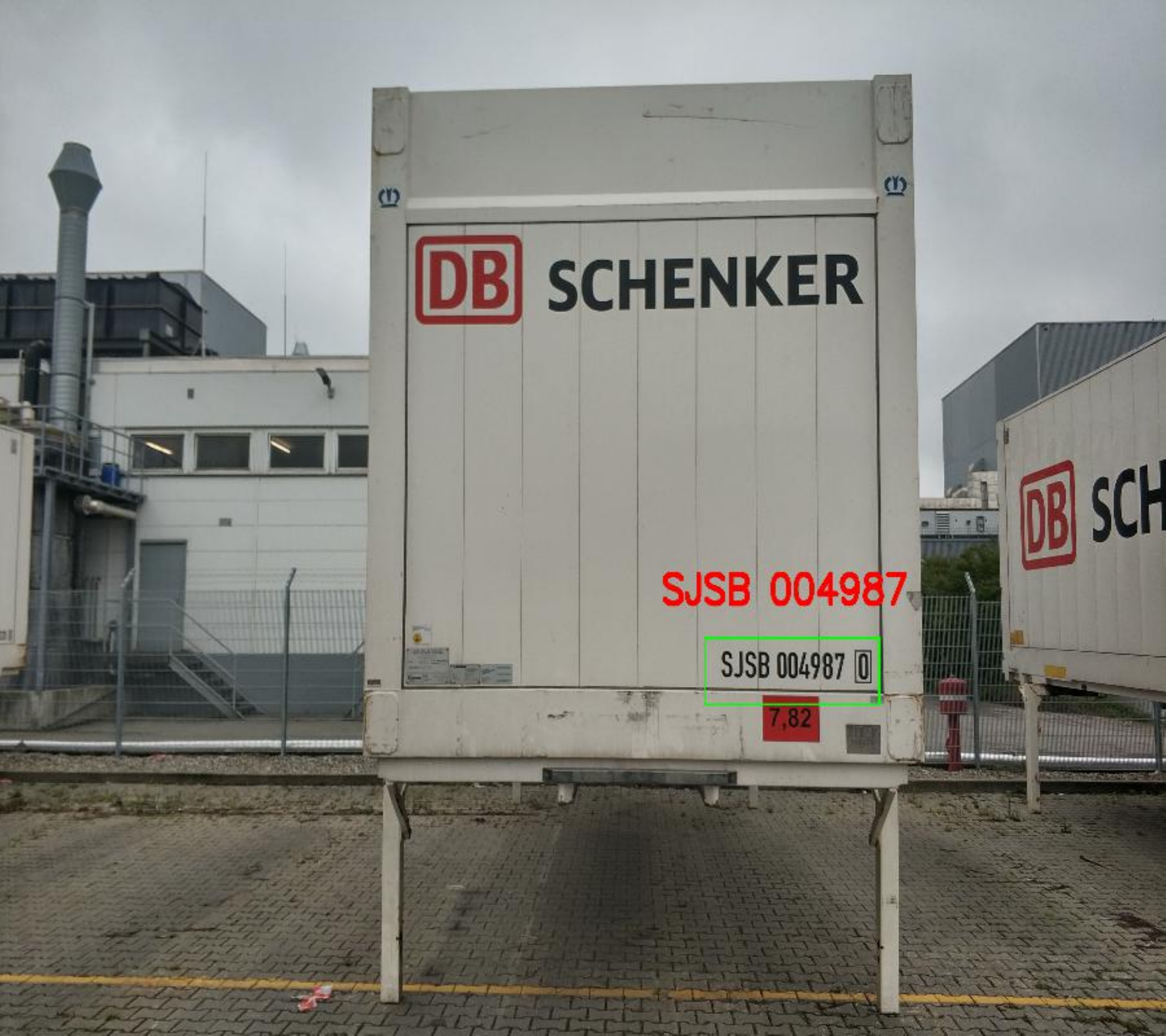}
    \caption{Output of ILU code recognition using Tesseract V4.}
  \end{minipage}
\end{figure}

\section{Swap-Body detection and ILU code reading}

The final system involves integrating the object detection model with the OCR pipeline. During testing with images, detection and identifying the type of swap-body by reading the ILU code using the OCR pipeline worked flawlessly. However, some assumptions where considered. When tested with images only the lower half of the detected bounding box was used for detecting the text. This approach ensured that only the ILU code area was considered for reading the text and improved the accuracy and robustness of the system. Figure 7 represents the final test results with images in various scenarios.

However, the implementation of object detection with OCR pipeline was a challenging task when tested with videos. Even though Tesseract V4 works perfectly under controlled situations, a deeper look into it revealed that the result can vary a lot with the presence of noise or if not properly pre-processed \cite{ramesh2018improving}. Moreover, the complexity of backgrounds, continuous variations in perspectives and viewpoints in a video are also contributing factors to the low performance of the OCR. For instance, one of the problems was in detecting number {\lq{5}\rq} which falsely detected as letter {\lq{S}\rq} in some viewpoints. Improving the quality of the input frame and skew correction were the initial attempts to correct this. In order to improve the quality of the input frame to the OCR pipeline, some additional preprocessing were performed. The aspect ratio of detected bounding boxes was continuously monitored and an aspect ratio threshold of 1.5 was used, below which the detected bounding boxes are not considered for text detection. This significantly, improved the results ensuring a minimum quality of input frame to the OCR pipeline. Finally, the object detector with OCR was tested successfully on videos without errors.

\begin{figure}[H]
\centering
\includegraphics[width=0.9\linewidth]{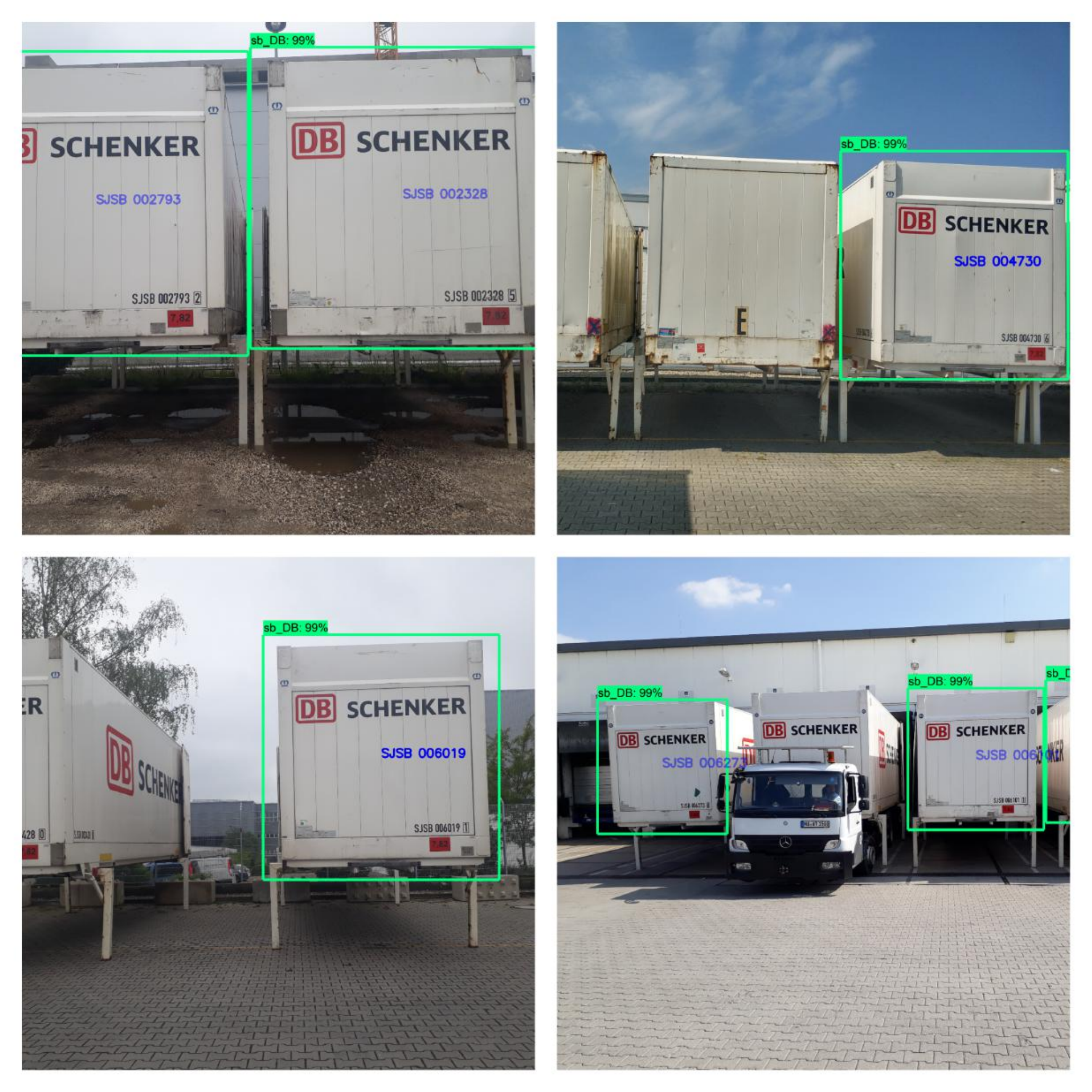}
\caption{Test results for object detection and OCR pipeline.}
\label{fig:verticalcell}
\end{figure}

\section{Conclusion}

Object detection is a crucial requirement concerning safety in operation for autonomous vehicles. An error can be vital and hence the object detection model should be accurate and robust. The model trained using the SSD algorithm performed excellently. Dataset collection and annotation was critical in determining the performance of the model. It was observed that incorporating images with complex scenarios including other objects as well for the training improved the performance of the model. Furthermore, high-quality images in good lighting conditions were used and more data augmentation techniques were applied. This significantly improved the overall mAP of the model. Optical character recognition gained good results when tested with images. Although text detection using EAST detector performed well, there were problems while identifying the text on videos.  After careful analysis, some of the shortcomings of Tesseract V4 in reading scene text was corrected. Continuous changes in viewpoint in videos resulted in false outputs when tested. Preprocessing the detected text bounding box resulted in significant improvements. In addition, improving the quality of video used for testing was also critical in achieving results without errors.

Even though implementing OCR with images were performing well, text recognition was not flawless when tested with videos. As observed, small variations in the input viewpoints resulted in large errors in the predicted text using Tesseract V4. Text recognition in a natural image is a difficult task. There are a number of ways to improve the performance of the OCR used in this project. Comparing the results with a separately trained classifier, such as SVM (Support Vector Machine) for detecting the ILU code can be one way to achieve this. As the next step to object detection, object tracking can be performed. It can also be a  possible solution to detect the ILU code correctly when the vehicle is at a reasonable distance from the swap-body.

%

%
%
\bibliographystyle{plain}
\bibliography{new_references}

\end{document}